\documentclass{article}

\usepackage[main, final]{neurips_2026}

\usepackage[utf8]{inputenc} % allow utf-8 input
\usepackage[T1]{fontenc}    % use 8-bit T1 fonts
\usepackage{hyperref}       % hyperlinks
\usepackage{url}            % simple URL typesetting
\usepackage{booktabs}       % professional-quality tables
\usepackage{amsfonts}       % blackboard math symbols
\usepackage{nicefrac}       % compact symbols for 1/2, etc.
\usepackage{microtype}      % microtypography
\usepackage{xcolor}         % colors
\usepackage{graphicx}
\usepackage{amsmath}
\usepackage{multirow}
\usepackage[table]{xcolor}
\usepackage{caption}
\usepackage{enumitem}

\newcommand{\hvs}{\textbf{\textit{HVS}}}
\newcommand{\hos}{\textbf{\textit{HOS}}}
\newcommand{\hps}{\textbf{\textit{HPS}}}

\title{Beyond Thinking: Imagining in 360$^\circ$ for\\Humanoid Visual Search}

\author{%
  Jingdong Zhang$^{1}$ \quad
  Yizhou Wang$^{2}$ \quad
  Zhengzhong Tu$^{1}$ \quad
  Xin Li$^{1}$ \\
  \bfseries
  Wenping Wang$^{1}$ \quad
  Xiaohang Zhan$^{2}$\thanks{Corresponding author.} \\[1.5ex]
  \normalfont
  $^1$Texas A\&M University \quad
  $^2$Adobe \\[1ex]
  \texttt{jdzhang@tamu.edu},\quad\texttt{xiaohangzhan@outlook.com}
}

\begin{document}
\maketitle

\begin{figure}[h]
  \centering
  \includegraphics[width=1.0\textwidth]{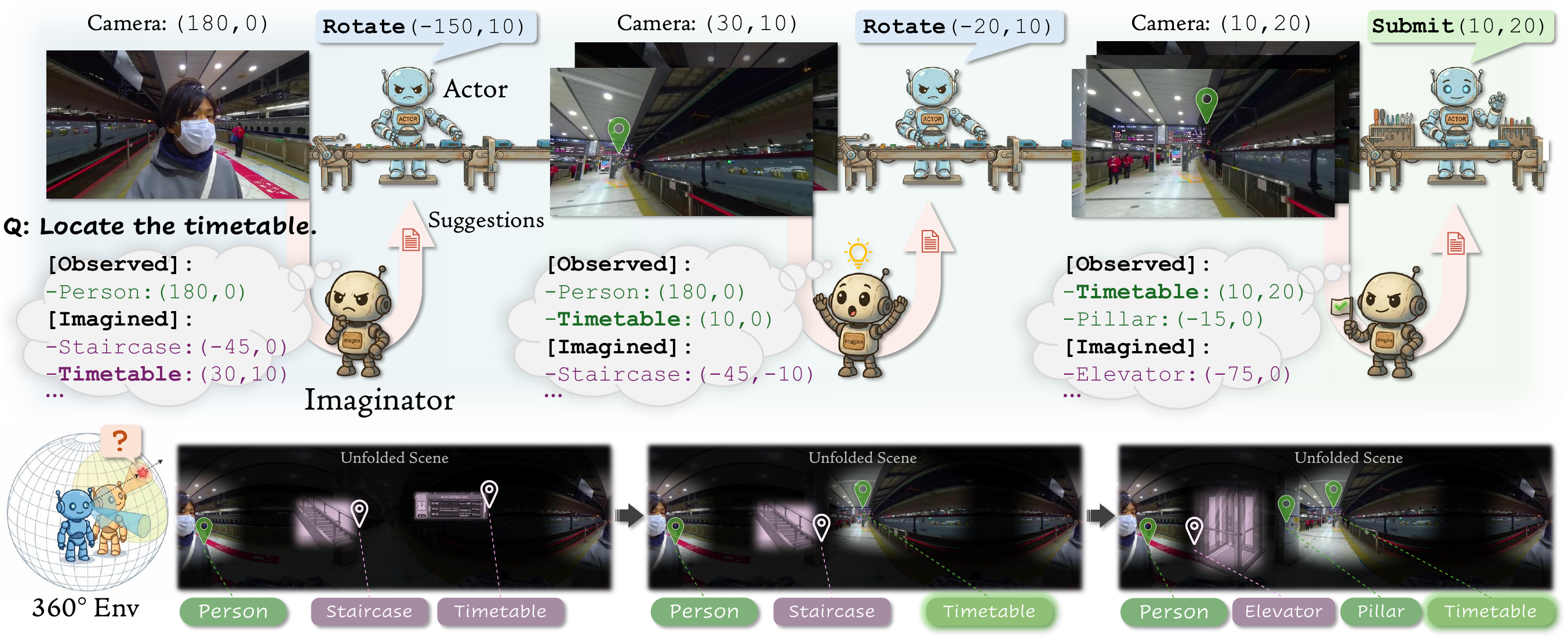}
  \caption{\textbf{Imagining in 360$^\circ$}: We propose a decoupled architecture for Humanoid Visual Search (\hvs). The \textbf{Imaginator} explicitly models the 360$^\circ$ environment by predicting the semantic layout of both \textit{observed} and \textit{unseen} regions, which provides the downstream \textbf{Actor} with a sampled distribution of spatial priors as suggestions. This explicit imagination empowers the agent to reorient its head more efficiently, hedging against spatial uncertainty in complex, in-the-wild environments.}
  \label{fig:teaser}
\end{figure}

\begin{abstract}
Humanoid Visual Search (\hvs) requires agents to actively explore immersive 360$^\circ$ environments. While prior methods treat this as a monolithic task relying on cumulative, multi-turn Chain-of-Thought (CoT) reasoning, they impose heavy cognitive burdens and require expensive trajectory-level annotations. In this paper, we propose \textbf{Imagining in 360$^\circ$}, a novel framework that decouples the exploration process into a specialized \textbf{Imaginator} and an \textbf{Actor}. The Imaginator functions as a probabilistic predictor of spatial priors; instead of maintaining a cumulative reasoning chain, it infers the semantic layout of both observed and unobserved regions in a single step. By sampling multiple hypotheses within this semantic space, we provide the Actor with a distribution of effective spatial information, offering robust guidance that hedges against uncertainty during active search. This decoupled architecture significantly lowers data engineering costs by eliminating the need for full-trajectory CoT annotations, enabling the generation of over 1.96 million curated training samples. Extensive experiments demonstrate that explicitly modeling semantic spatial priors drastically improves search efficiency and success rates in complex, in-the-wild environments.
\end{abstract}

\section{Introduction}
\label{intro}

The human visual system efficiently navigates complex spaces by actively reorienting the head and eyes to deliberately uncover new visual information~\cite{marr2010vision,stein2024eye}. Replicating this behavior in artificial agents has driven the development of \textbf{Humanoid Visual Search (\hvs)}, an embodied task where agents must actively explore immersive 360$^\circ$ panoramic environments to locate specific objects (\hos) or navigable paths (\hps). Unlike passive perception, \hvs\ requires a closed-loop perception-action cycle~\cite{torralba2006contextual}. To tackle this, a pioneering work, \textit{Thinking in 360$^\circ$}~\cite{yu2025thinking}, formulates the search as a sequential decision-making process powered by Multimodal Large Language Models (MLLMs)~\cite{wu2024v}. Their approach employs a single-model, Chain-of-Thought (CoT)~\cite{Wei2022ChainOT} framework where the agent iteratively accumulates narrow field-of-view observations, maintains a historical memory of the explored space, and plans the next head-turning action within a unified reasoning chain.

However, this monolithic paradigm introduces significant bottlenecks. First, humanoid visual reasoning in 3D environments remains highly challenging for current MLLMs~\cite{yu2025thinking}, with the coupling the challenging spatial understanding with the execution of action planning into a single model. Second, because the model lacks the explicit modeling of spatial relationships, it is forced to implicitly guess the layout of unseen regions, leading to inactive or inaccurate explorations. Finally, teaching the MLLM as a single agent for this process relies heavily on curating high-quality, long-context CoT reasoning trajectories. The prohibitive cost of acquiring such dense, multi-step expert annotations fundamentally limits the ability to scale these systems and hinders their generalization across diverse, in-the-wild scenarios.

In this work, we propose a fundamental shift in how embodied agents approach unobserved space, effectively redefining the \hvs\ paradigm. Our core insight is that human-like search is not strictly driven by exhaustive logical chains alone, but is heavily guided by intuitive spatial priors (\textit{e.g.} expecting an open corridor likely leads to an entrance or exit). Thus, we propose \textbf{Imagining in 360$^\circ$}, a framework designed to explicitly decouple probabilistic spatial imagination from rigorous CoT reasoning. By introducing intuitive spatial priors to guide the cumulative reasoning process, we transform the exploration task into a collaboration between probabilistic predicting of spatial priors and rigorous reasoning, which encourages the agent to explore more actively and confidently, grounded in sound spatial assumptions.

To achieve this, we introduce an architecture consisting of a decoupled \textbf{Imaginator} and an \textbf{Actor}. Rather than relying on computationally heavy pixel-level reconstruction~\cite{he2022masked} or visual inpainting~\cite{lugmayr2022repaint} to guess unseen areas, we propose to efficiently model the visual horizon directly within the semantic space of language. Given limited narrow field-of-view (NFoV) observations, the Imaginator predicts the layout of both \textit{observed} and \textit{imagined} coordinates of highly semantic regions. Crucially, the Imaginator functions as a probabilistic predictor of spatial priors, instead of outputting a single deterministic guess, it generates multiple spatial hypotheses per step via temperature sampling to capture the inherent uncertainty of occluded environments. As shown in Fig.~\ref{fig:teaser}, these semantic layouts are processed as a ranked list of suggestions and explicitly provided to the downstream Actor's context. By evaluating this probabilistic distribution, the Actor gains comprehensive spatial awareness before executing its next head rotation.

Beyond enhancing spatial reasoning, this decoupling unlocks massive data scalability. Because the Imaginator learns spatial priors directly from single-step semantic annotations rather than cumulative CoT histories, we can introduce a fully automated pseudo-label pipeline. This approach effortlessly transforms unlabeled 360$^\circ$ panoramas into high-quality training data, scaling our dataset to $\sim$1.92 million samples across $\sim$20k scenes without manually annotated CoT trajectories.

In summary, our main contributions are as follows:
\begin{itemize}
    \item \textbf{Decoupled HVS Paradigm:} We redefine visual search by separating intuitive spatial imagination from action planning, allowing for more active and grounded exploration.
    \item \textbf{Probabilistic Imaginator:} We introduce an Imaginator model that learns intuitive spatial priors and guides the Actor agent with probabilistic estimations.
    \item \textbf{Scalable Data Engine:} Our approach enables a fully automated pipeline that scales training to 1.92M samples without manual trajectory labels, ensuring robust performance in the wild.
\end{itemize}

\section{Related Works}
\label{related}
\vspace{-2mm}
\noindent \textbf{Embodied Visual Search.} 
Traditional visual search relies on visual saliency or contextual cues~\cite{oliva2003top, torralba2006contextual, zelinsky2005role}. While Multimodal Large Language Models (MLLMs) like \textit{V$^*$}~\cite{wu2024v} have advanced these capabilities, they predominantly operate on static, disembodied 2D images. However, real-world search is an active, embodied behavior requiring coordinated head and eye movements to uncover occluded information~\cite{lanman1978coordination, barnes1979vestibulo, solman2017eye}. Emulating this has driven the emergence of Humanoid Visual Search (\hvs)~\cite{yu2025thinking}, where agents actively explore immersive 360$^\circ$ spaces. While pioneering works~\cite{yu2025thinking} tackle \hvs via cumulative Chain-of-Thought (CoT) reasoning, this monolithic paradigm imposes a severe cognitive burden and limits data scalability. To overcome this, our framework explicitly decouples spatial imagination from action.

% Traditionally, visual search has been formulated as a deliberate cognitive process for identifying targets within cluttered scenes, historically relying on bottom-up visual saliency or top-down contextual cues~\cite{oliva2003top, torralba2006contextual, zelinsky2005role}. While recent breakthroughs powered by Multimodal Large Language Models (MLLMs), such as \textit{V$^*$}~\cite{wu2024v}, have significantly advanced search capabilities by leveraging rich world knowledge, these methods predominantly evaluate perception on static, disembodied 2D images. However, visual search in the physical world is inherently an active and embodied behavior. Humans naturally coordinate head and eye movements to uncover occluded information in complex environments~\cite{lanman1978coordination, barnes1979vestibulo, solman2017eye}. Emulating this active perception has driven the emergence of Humanoid Visual Search (\hvs)~\cite{yu2025thinking}, requiring agents to explore immersive 360$^\circ$ spaces within a closed-loop perception-action cycle. While pioneering works like~\cite{yu2025thinking} approach \hvs via cumulative Chain-of-Thought (CoT) reasoning, this monolithic paradigm imposes a severe cognitive burden on the model and restricts data scalability. To overcome this, our approach explicitly decouples spatial imagination from action.

\noindent \textbf{Agentic Multimodal LLMs.}
Multimodal LLMs (MLLMs) represent a significant stride toward unified understanding models among multi-modalities~\cite{alayrac2022flamingo,liu2023llava,li2022blip,lai2024lisa,qwen3}. To transcend their innate perceptual limitations, recent MLLM agents have been empowered with external toolkits (\textit{e.g.}, web browsing, code execution) and refined via multi-turn reinforcement learning~\cite{guo2025deepseek, jin2025search, team2025kimik15}. In the visual domain, agents leverage symbolic tool calls like iterative zooming or region selection to achieve fine-grained understanding~\cite{li2025dyfo, shen2024zoomeye, zhang2025chain,wang2025pixel}. Despite these advancements, these tool-use paradigms often occur on a disembodied 2D canvas. While some recent efforts ground MLLMs in embodied reasoning for robotics~\cite{azzolini2025cosmos, team2025gemini, Chen25-ecot-lite}, active visual search driven by a generative spatial prior remains largely underexplored. We introduce an \textit{Imaginator} as a specialized agentic component that provides probabilistic spatial guidance, enabling more efficient and robust search behaviors.

\noindent \textbf{Spatial Imagination and Scene Extrapolation.} Effective embodied exploration requires anticipating regions beyond the limited narrow field-of-view (NFoV). Prior methods typically rely on pixel-level extrapolation~\cite{shi2023mvdream, Tang2023mvdiffusion, huang2025vista, huang2025vistav2, he2022masked, zhang2025spgen}, which is computationally expensive and artifact-prone, or implicit latent spaces~\cite{lecun2022path, assran2023self, bardes2024revisiting,huang2025llm}, which lack the explicit semantic grounding required for language-driven planning. Meanwhile, out-of-view MLLMs~\cite{chen2025openview} and scene graphs~\cite{gu2024conceptgraphs} often restrict themselves to simple VQA or demand exhaustive multi-view scanning. Departing from these, we frame spatial imagination as generating \textit{probabilistic spatial priors}. Inspired by biological cognitive mapping~\cite{marr2010vision, bartz1966eye, tolman1948cognitive}, our \textit{Imaginator} extrapolates the 360$^\circ$ topology entirely in language space. By predicting coordinates for visible and occluded entities, we bypass pixel-level overhead and CoT data curation, offering interpretable and efficient guidance for navigation.

\begin{figure}[t]
  \centering
  \includegraphics[width=1.0\textwidth]{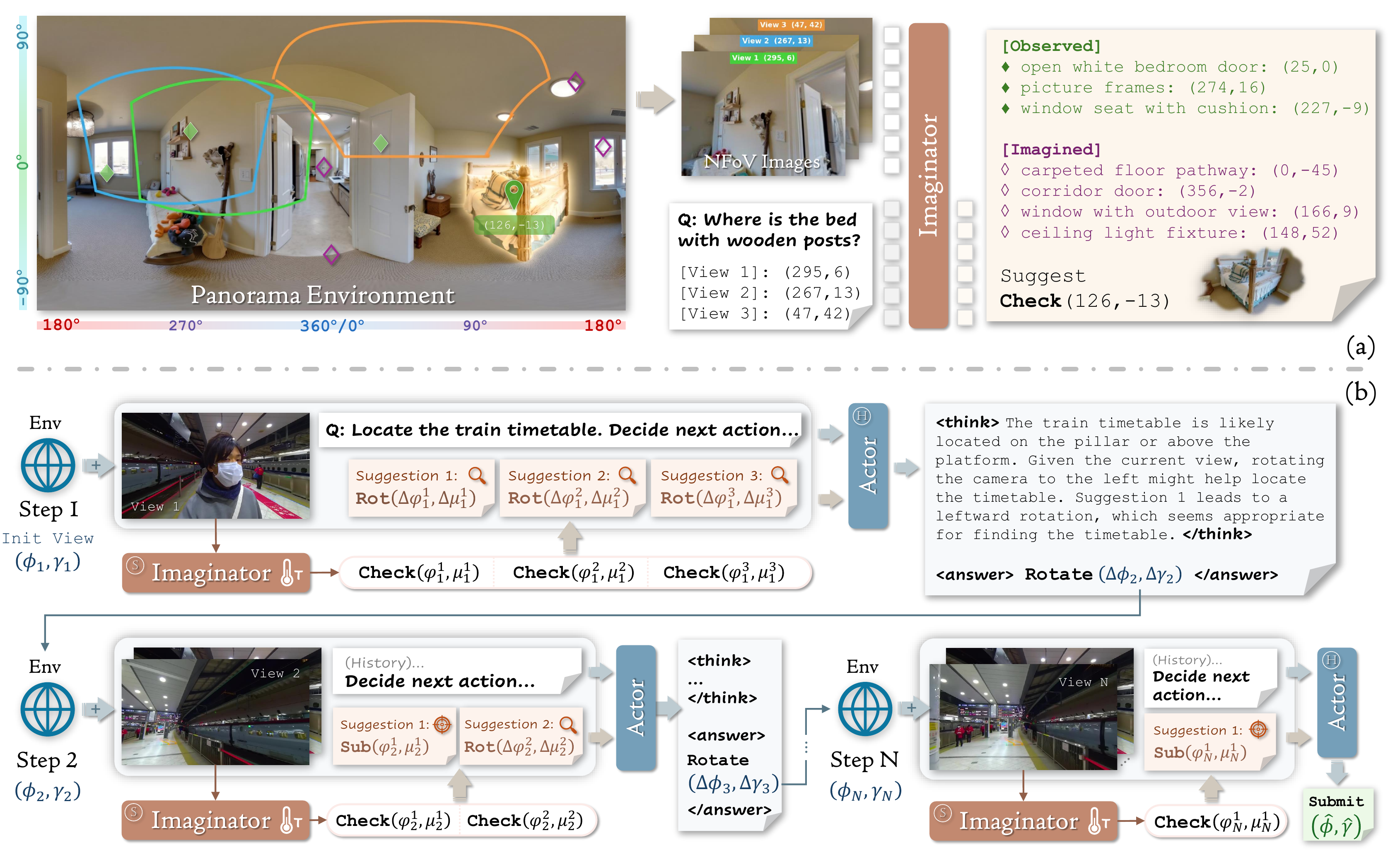}
   \vspace{-4mm}
    \caption{\textbf{Framework Overview.} 
  (a) \textbf{360$^{\circ}$ Env Imagination}: The Imaginator performs a single-step estimation of the global layout by predicting $(\varphi, \mu)$ coordinates for both observed landmarks and imagined entities to capture high-level spatial relationships. 
  (b) \textbf{Collaborative HVS Pipeline}: The Actor integrates the sampled distribution of probabilistic spatial suggestions into its reasoning chain, allowing the agent to hedge against environmental uncertainty and guide active search efficiently.}
  \vspace{-3mm}
\label{fig:main}
\end{figure}

\vspace{-2mm}
\section{Method}
\label{sec:method}
\vspace{-2mm}
To overcome the cognitive overload and data-scalability bottlenecks inherent in monolithic CoT search paradigms, we propose a decoupled architecture for Humanoid Visual Search (\hvs). As illustrated in Fig.~\ref{fig:main}, our framework separates the active search process into two distinct roles: an \textbf{Imaginator} that explicitly models the 360$^\circ$ environment and generates probabilistic spatial priors, and an \textbf{Actor} that acts upon these priors to navigate in it.

\vspace{-1mm}
\subsection{Problem Formulation}
\label{subsec:problem}
\vspace{-1mm}
\paragraph{Objective.} We formulate the environment as a continuous 360$^\circ$ panorama. Because the embodied agent possesses a limited narrow field-of-view (NFoV), it must actively rotate its head to explore the surrounding space. At any step, the agent receives a NFoV observation $o_{\phi,\gamma} \in \mathcal{O}$, defined by its current azimuth ($\phi$) and polar angle ($\gamma$). Given a natural language instruction $x$, the overarching goal of \hvs\ is to iteratively gather visual evidence and ultimately predict a final viewing direction $(\hat{\phi}, \hat{\gamma})$ that maximizes the probability of task success $r_s$:
\begin{equation}
(\hat{\phi}, \hat{\gamma}) = \arg \max_{\phi,\gamma} P(r_s \mid o_{\phi,\gamma}, x).
\end{equation}
\vspace{-1mm}
Following the task definition in \cite{yu2025thinking}, \textbf{Humanoid Object Search (HOS)} requires the agent to locate a specific target and submit a final view $(\hat{\phi}, \hat{\gamma})$ that locates the object, while \textbf{Humanoid Path Search (HPS)} involves identifying a final orientation $\hat{\phi}$ aligned with a navigable path.

\subsection{The Imaginator Model}
\label{subsec:imagination}
\vspace{-1mm}
Rather than forcing a single agent to implicitly guess the layout of unobserved areas, we introduce the \textbf{Imaginator}. Its sole purpose is to construct a semantic layout map of both observed and unobserved regions based on the current visual history, as shown in~\ref{fig:main} (a), acting as a generative spatial prior.
\vspace{-1mm}
\paragraph{Task Setup and Model Execution.} 
Spatial object layouts in the physical world strictly adhere to inherent geometric and semantic rules (\textit{e.g.}, a chandelier typically hangs from the ceiling rather than sitting on the floor; a teacup rests on a table rather than a wall). Internalizing these structural priors is crucial for accurate \hvs, as they drastically narrow down redundant explorations and facilitate fast convergence. Therefore, our primary objective is to enable the model to learn these spatial priors from the data distribution to robustly guide the search process.

A naive approach might treat this as a direct target detection task, simply mapping the instruction $x$ and historical observations to a target coordinate $(\varphi, \mu)$. However, due to the severely limited field-of-view, this direct mapping often causes the solution space to collapse into a sharply peaked distribution. Deprived of sufficient visual context, the model is highly susceptible to generating high-confidence but erroneous predictions. 

To foster reasonable exploratory behaviors akin to human intuition, we design the Imaginator to perform a comprehensive layout estimation rather than a solitary point prediction. At step $t$, receiving all visible FoV images $\{o_{\phi_1,\gamma_1}, \dots, o_{\phi_t,\gamma_t}\}$ and the instruction $x$, the Imaginator executes a stateless, single-step forward pass. It outputs a structured block that explicitly lists semantic entities and their predicted absolute coordinates $(\varphi, \mu)$ in the equirectangular space:
\begin{itemize}
    \item \texttt{[Observed]}: An estimation of landmarks currently or previously visible in the NFoVs (\textit{e.g.}, ``\textit{picture frames: (274, 16)}'').
    \item \texttt{[Imagined]}: A probabilistic prediction of objects likely existing in \textit{unseen} regions (\textit{e.g.}, ``\textit{window with outdoor view: (166, 9)}'').
\end{itemize}
By enforcing this joint prediction of both observed and imagined entities, the Imaginator explicitly models the spatial relationships of key visual anchors. This structural grounding makes it significantly easier to infer the target's location, naturally producing a smoother distribution supporting the sampling of spatial priors. The final output is a brief recommendation \texttt{Suggest Check $(\varphi, \mu)$}.

% Crucially, compared to previous world/imagine models that rely on pixel-space reconstruction or visual inpainting \cite{huang2025vista}, which inevitably squander computational bandwidth on modeling complex, low-level real-world noise~\cite{lecun2022path, assran2023self}, our approach operates entirely within a highly semantic coordinate space. This abstraction effectively isolates the essential spatial relationships, offering a more efficient mechanism for world modeling.

\begin{figure}[t]
  \centering
  \includegraphics[width=1.0\textwidth]{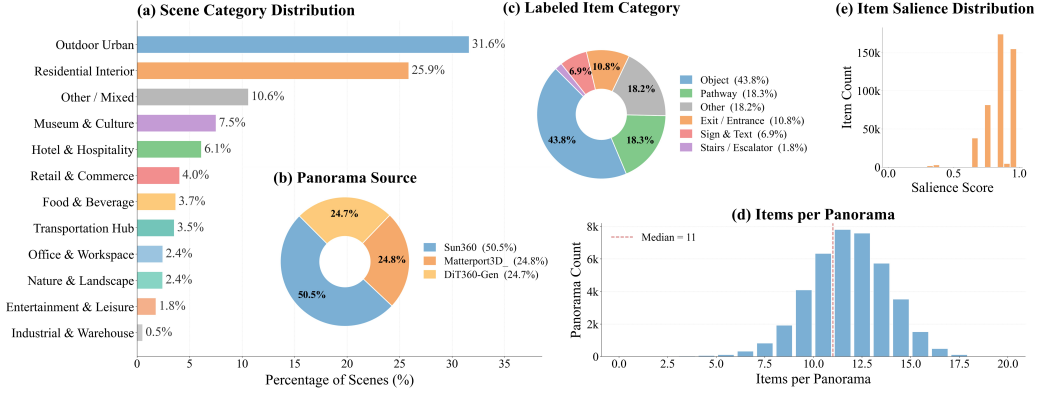}
   \vspace{-4mm}
  \caption{\textbf{Overview of the Curated Training Dataset.} (a) Distribution of scene categories, highlighting a diverse mix dominated by most commonly encountered outdoor urban (31.6\%) and residential interior (25.9\%) environments. (b) The composition of unlabeled panorama sources. (c) The distribution of labeled semantic item categories. (d) The histogram of labeled items per panorama, with a median of 11 items, demonstrates dense semantic coverage for spatial modeling. (e) Salience distribution of the labeled items in the environment.}
  \vspace{-3mm}
  \label{fig:data}
\end{figure}

\paragraph{Automated Scale-Up Data Engine.} 
Training the Imaginator is inherently more scalable than prior CoT-based methods~\cite{yu2025thinking}. Instead of relying on teacher models for long, multi-step trajectories that require costly manual curation, our decoupled architecture frames spatial learning simply as a \textit{masked spatial modeling} task on panoramas. Specifically, we utilize a powerful teacher model to perform accurate and comprehensive labeling across the complete 360$^\circ$ environment. From these fully annotated panoramas, we synthetically sample trajectories and reveal only 1 to 8 NFoV observations per training instance (with 50$\%$ targets kept out of views to force spatial layout imagination). By feeding the Imaginator these limited views and tasking it to predict the entire layout (following the \texttt{[Observed]} and \texttt{[Imagined]} format discussed above), we actively force the model to comprehend the global environment and internalize spatial relationships. This paradigm completely bypasses the prohibitive cost of trajectory-level CoT annotations, enabling a fully automated, low-cost pseudo-labeling pipeline. As summarized in Fig.~\ref{fig:data}, we scale our training set using approximately 20k diverse panoramic environments filtered from~\cite{zhang2026mtpano}, including real-world and synthetic subsets. Through deterministic trajectory expansion, this efficiently yields $\sim$1.92 million high-quality, single-step spatial reasoning samples.

\paragraph{Training Objective.}
Formally, given the instruction $x$ and visual history $\mathcal{O}_t = \{o_{\phi_1,\gamma_1}, \dots, o_{\phi_t,\gamma_t}\}$, the Imaginator generates the target sequence $y = \{y_1, \dots, y_L\}$, comprising the structured layouts and spatial suggestions. We optimize the model parameter $\theta$ using standard auto-regressive cross-entropy loss applied only to the completion tokens:
\vspace{-1mm}
\begin{equation}
\mathcal{L} = - \sum_{i=1}^{L} \log P_\theta(y_i \mid x, \mathcal{O}_t, y_{<i})
\end{equation}
To efficiently internalize 3D spatial priors, the model is first pretrained on the large-scale synthetic data, and subsequently finetuned on the processed \hvs\ training split in~\cite{yu2025thinking} with the same format for precise task alignment.

\subsection{The Actor Model}
\label{subsec:actor}
\vspace{-1mm}
The \textbf{Actor} is the agent responsible for executing the physical search within the environment. Our framework is agnostic to the specific architecture of the Actor; it can be any off-the-shelf MLLM or a fine-tuned search model, \textit{e.g.}, the HVS-3B used in~\cite {yu2025thinking}. Following their standard \hvs\ formulation, the \textbf{Actor} is a tool-augmented model responsible for executing the physical search. At step $t$, the Actor receives the target instruction $x$, the current observation $o_{\phi_t,\gamma_t}$, its historical memory, and crucially, the spatial suggestions $\mathcal{S}_t$ provided by the Imaginator. 

Empowered by these probabilistic spatial priors, the Actor generates a textual reasoning chain (the \texttt{<think>} block) to assess the plausibility of the suggested layouts against its own visual memory. It concludes by outputting an \texttt{<answer>} block specifying the next action from two primitives:
\begin{itemize}[leftmargin=*, nosep]
    \item \textbf{\texttt{Rot}} $(\Delta\phi_{t+1}, \Delta\gamma_{t+1})$: Adjusts the viewing direction relative to the current camera pose.
    \item \textbf{\texttt{Sub}} $(\hat{\phi}, \hat{\gamma})$: Commits the current view as the final target estimate and terminates the search.
\end{itemize}

\subsection{Imagination-Guided HVS Pipeline}
\label{subsec:pipeline}
\vspace{-1mm}
While the Imaginator provides scalable spatial priors, its stochastic predictions are unsuitable for final navigation decisions. Conversely, a standalone Actor lacks explicit spatial modeling and often hallucinates in unseen regions. We thus propose a complementary pipeline: the Imaginator generates probabilistic priors, which the Actor grounds via rigorous sequential reasoning.

As shown in Fig.~\ref{fig:main} (b), to fully exploit the Imaginator's spatial knowledge without over-relying on a single potentially flawed layout, we formulate its output as a probabilistic distribution. At each step $t$, we sample $K$ candidate coordinates, $\mathcal{C}_t = \{(\varphi_t^1, \mu_t^1), \dots, (\varphi_t^K, \mu_t^K)\}$, extracted from its \texttt{Suggest Check $(\varphi, \mu)$} outputs. The first candidate is generated via greedy decoding ($T=0$) to anchor the most confident prediction. The remaining $K-1$ candidates are stochastically sampled ($T>0$) to capture diverse spatial hypotheses. The temperature $T$ follows a step-wise decay schedule over the search steps: as the agent accumulates more visual clues, the environmental uncertainty naturally decreases, transitioning the sampling strategy from broad exploration to focused prediction.

Before injecting the hypotheses into the Actor's context, they are translated into actionable primitives relative to the Actor's current camera pose $(\phi_t, \gamma_t)$. We deploy a suggestion converter to process each absolute predicted coordinate $(\varphi_t^i, \mu_t^i)$. First, it calculates the required angular displacement:
\begin{equation}
\Delta\varphi_t^i = (\varphi_t^i - \phi_t) \bmod 360^\circ, \quad \Delta\mu_t^i = \mu_t^i - \gamma_t
\end{equation}
Next, the converter evaluates whether the imagined target location already falls within the boundaries of the current NFoV, denoted by angular bounds $(F_\phi, F_\gamma)$. If the coordinate is within the current view, further exploration is unnecessary, and the suggestion is mapped to a termination action. Otherwise, it is mapped to a rotation action directing the agent toward the unobserved region. Formally, the mapped suggestion $s_t^i$ is defined as:
\begin{equation}
s_t^i = 
\begin{cases} 
\texttt{Sub}(\varphi_t^i, \mu_t^i), & \text{if } |\Delta\varphi_t^i| \le \frac{F_\phi}{2} \text{ and } |\Delta\mu_t^i| \le \frac{F_\gamma}{2} \\
\texttt{Rot}(\Delta\varphi_t^i, \Delta\mu_t^i), & \text{otherwise}
\end{cases}
\end{equation}
This mapping yields a final, ranked set of actionable suggestions $\mathcal{S}_t = \{s_t^1, s_t^2, \dots, s_t^K\}$. These suggestions are seamlessly injected into the Actor's prompt under the specific tag \texttt{[Spatial Imagination Suggestions]}. By explicitly evaluating this distribution of spatial priors, the Actor obtains awareness of the spatial layout, and significantly suppressing overconfident exploration errors and accelerating convergence in complex environments.

\begin{table}[t]
\centering
\small
\caption{\textbf{Results on H*Bench.} We report the Success Rate (\%) across various Vision-Language Model families. `$\times$' denotes the baseline Actor relying solely on its internal CoT. `$\checkmark$' indicates the Actor plugged in with our Imaginator's spatial priors.}
\setlength{\tabcolsep}{2.3mm}{\scalebox{0.75}{
\begin{tabular}{llcccccccccc}
\toprule
\multicolumn{2}{c}{Actor} & \multirow{2}{*}{Imaginator} & \multicolumn{4}{c}{Humanoid Object Search} & \multicolumn{5}{c}{Humanoid Path Search} \\
\cmidrule(lr){1-2} \cmidrule(lr){4-7} \cmidrule(lr){8-12}
Family & Params &  & \textbf{Overall} & Easy & Medium & Hard & \textbf{Overall} & Easy & Medium & Hard & Extreme \\
\midrule
\rowcolor{gray!15} \multicolumn{12}{l}{\textit{Open-Source Models}} \\
\multirow{6}{*}{Qwen2.5-VL} & \multirow{2}{*}{3B} & $\times$ & 21.17 & 18.54 &  16.48 &  22.68  & 7.69 & 9.20 & 9.65 & 3.50 & 8.33 \\
 &  & $\checkmark$ & 49.54 & 59.19 &  51.70 &  45.62 & 25.94 & 34.80 & 22.15 & 24.77 & 15.74 \\
\cmidrule(lr){2-12}
 & \multirow{2}{*}{7B} & $\times$ & 12.79 & 31.22 & 7.39 & 6.34 & 9.88 & 12.20 & 11.18 & 7.24 & 6.94 \\
 &  & $\checkmark$ & 57.42 & 70.73 & 51.70 & 52.95 & 28.38 & 35.20 & 24.34 & 29.44 & 18.98 \\
\cmidrule(lr){2-12}
 & \multirow{2}{*}{72B} & $\times$ & 29.54 & 36.91 & 21.59 & 27.59 & 12.94 & 17.40 & 9.21 & 15.65 & 5.09 \\
 &  & $\checkmark$ & 62.25 & 70.08 & 60.80 & 59.42 & 38.19 & 45.00 & 36.18 & 39.25 & 24.54 \\
\midrule
\multirow{4}{*}{Qwen3-VL} & \multirow{2}{*}{32B} & $\times$ & 34.38 & 66.83 & 36.36 & 21.75 & 23.19 & 26.00 & 24.78 & 22.43 & 14.81 \\
 &  & $\checkmark$ & 57.46 & 73.98 & 51.70 & 51.77 & 31.31 & 37.60 & 30.04 & 30.14 & 21.76 \\
\cmidrule(lr){2-12}
 & \multirow{2}{*}{235B} & $\times$ & 23.62 & 57.24 & 26.70 & 10.44 & 21.44 & 26.80 & 19.96 & 23.36 & 8.33 \\
 &  & $\checkmark$ & 65.04 & 81.46 & 60.23 & 59.29 & 38.25 & 46.60 & 35.31 & 37.15 & 27.31 \\
\midrule
\multirow{4}{*}{Gemma-3} & \multirow{2}{*}{4B} & $\times$ & 18.17 & 41.79 & 25.57 & 8.33 & 14.50 & 17.80 & 16.67 & 10.98 & 9.26 \\
 &  & $\checkmark$ & 57.75 & 72.03 & 60.23 & 52.02 & 35.06 & 43.80 & 27.41 & 36.45 & 28.24 \\
\cmidrule(lr){2-12}
 & \multirow{2}{*}{12B} & $\times$ & 22.88 & 54.96 & 32.39 & 9.57 & 15.31 & 16.80 & 16.23 & 14.49 & 11.57 \\
 &  & $\checkmark$ & 54.62 & 63.90 & 48.86 & 51.71 & 33.56 & 43.00 & 30.04 & 32.24 & 21.76 \\
 \midrule
 \multirow{2}{*}{Kimi-VL} & \multirow{2}{*}{A3B} & $\times$ & 4.33 & 12.36 & 0.57 & 1.68 & 5.62 & 7.20 & 6.80 & 2.34 & 6.02 \\
 &  & $\checkmark$ & 32.46 & 40.00 & 27.84 & 30.08 & 21.00 & 28.20 & 16.89 & 21.96 & 11.11 \\
\midrule
 \rowcolor{gray!15} \multicolumn{12}{l}{\textit{Proprietary Models}} \\
 \multirow{4}{*}{Gemini-2.5} & \multirow{2}{*}{Flash} & $\times$ & 43.70 & 79.50 & 59.70 & 28.30 & 27.70 & 35.80 & 22.60 & 31.30 & 12.50 \\
 &  & $\checkmark$ & 63.20 & 86.30 & 69.30 & 53.70 & 37.70 & 50.20 & 30.30 & 42.30 & 15.30 \\
 \cmidrule(lr){2-12}
 & \multirow{2}{*}{Pro} & $\times$ & 55.00 & 88.00 & 62.50 & 41.60 &  37.90 & 43.80 & 36.40 & 40.70 & 21.80 \\
 &  & $\checkmark$ & {66.10} & 89.60 & 68.80 & 56.80 & {43.10} & 51.00 & 39.90 & 48.80 & 20.40 \\
 \midrule
 \rowcolor{gray!15} \multicolumn{12}{l}{\textit{Finetuned Models}} \\
 \multirow{2}{*}{HVS} & \multirow{2}{*}{3B} & $\times$ & 48.04 & 61.79 & 26.14 & 45.18 & 24.12 & 36.60 & 17.76 & 22.66 & 11.57 \\
 &  & $\checkmark$ & 62.75 & 75.77 & 64.20 & 57.61 & 39.38 & 49.40 & 32.89 & 41.12 & 26.39 \\
\bottomrule
\end{tabular}
}}
\label{tab:main_results}
\end{table}

\vspace{-2mm}
\section{Experiment}
\vspace{-1mm}
\subsection{Implementation Details}
\label{subsec:implementation}
\vspace{-1mm}
\paragraph{Data Curation.} To construct the pseudo-pretrained dataset, we employ Qwen3-VL-235B as the teacher model and to filter $\sim$20k valid indoor/outdoor scenes from Sun360~\cite{xiao2012recognizing}, Matterport3D~\cite{chang2017matterport3d}, and synthesized by DiT360~\cite{feng2025dit360}. Then we localize 5-15 salient semantic targets per panorama. For each scene, we synthetically render 24 NFoV trajectories, in which 50$\%$ of the targets are deliberately withheld from the field-of-view. Deterministically expanding these trajectories based on FoV coverage yields $\sim$1.92M single-step \texttt{[Observed]}/\texttt{[Imagined]} samples. We also curate 2,100 \hos, and 2,148 \hps imagine data with the same format on H*Bench \cite{yu2025thinking} for task alignment.
\vspace{-1mm}
\paragraph{Imaginator Training.} The Imaginator is initialized from Qwen3-VL-8B-Instruct~\cite{qwen3}. We first train the full model on 1.92M pseudo dataset for 1 epoch with a global batch size of 192 to generate structured imagination formats, instilling broad imagination priors across diverse scene types. Subsequently, we perform full-parameter supervised finetuning on the processed HVS training split to focus strictly on task alignment. This SFT runs for 2 epochs with a global batch size of 8 and a learning rate of $1 \times 10^{-5}$. The experiments are conducted on 8 NVIDIA H200/A100 GPUs.
\vspace{-1mm}
\paragraph{Actor and Inference.} Our framework is architecture-agnostic. We validate generalization across off-the-shelf models including \textit{open-source}: Qwen2.5-VL~\cite{xu2025qwen25vltechnicalreport}, Qwen3-VL~\cite{qwen3}, Gemma-3~\cite{gemmateam2025gemma3technicalreport}, and Kimi-VL~\cite{kimiteam2025kimivltechnicalreport}; \textit{proprietary models}: Gemini 2.5~\cite{sadigh2025gemini25pushing}; \textit{finetuned models:} HVS-3B~\cite{yu2025thinking}. During inference, the Imaginator generates three spatial hypotheses per step: one greedy anchor ($T=0$) and two stochastic samples ($k=50$, initial $T_1=0.7$ with a 0.85 step-wise decay). These absolute predictions are converted into ego-centric relative actions and appended to the Actor's context.
\vspace{-1mm}
\paragraph{Evaluation Protocol.} Following \cite{yu2025thinking}, search episodes are within 10 steps. NFoV observations are rendered at $960 \times 720$ resolution with a 100$^\circ$ FoV. Success requires the final submitted view to fall within the bounding-box-centered angular tolerance: $\tau_\phi=30^\circ$ and $\tau_\gamma=20^\circ$ for \hos, and $\tau_\phi=10^\circ$ for \hps.

\subsection{H*Bench Evaluation}
\label{subsec:main_results}

Table~\ref{tab:main_results} presents the comprehensive evaluation of our framework across a diverse collections of \textit{Open-Source}, \textit{Proprietary}, and domain-specific \textit{Finetuned} MLLMs. For off-the-shelf open-source models, the monolithic CoT approach (Actor only) generally struggles with the spatial reasoning, often resulting in overall success rates below 35\% on \hos, 25\% on \hvs. However, when augmented with the Imaginator, models like Qwen3-VL (235B) surge from 23.62\% to 65.04\% on \hos, and from 21.44\% to 38.25\% on \hps. Remarkably, this plug-and-play compatibility extends to powerful proprietary models; Gemini-2.5 Flash/Pro receives notable gain on both splits. Most importantly, our decoupled pipeline significantly outperforms the previous state-of-the-art established by RL-finetuned models. The baseline HVS-3B model achieves 48.04\% on \hos\ and 24.12\% on \hps. When integrated into our joint pipeline as the Actor, its performance jumps to 62.75\% and 39.38\%, respectively. This validates the limitation, the lack of explicit, probabilistic spatial modeling, in current single-agent architectures.

\begin{figure}[t]
  \centering
  % 左侧图片区域：占据 45% 的页面宽度
  \begin{minipage}[c]{0.52\textwidth}
    \centering
    % 注意：这里的宽度改为 \linewidth，它会自动适应当前 minipage 的宽度
    \includegraphics[width=\linewidth]{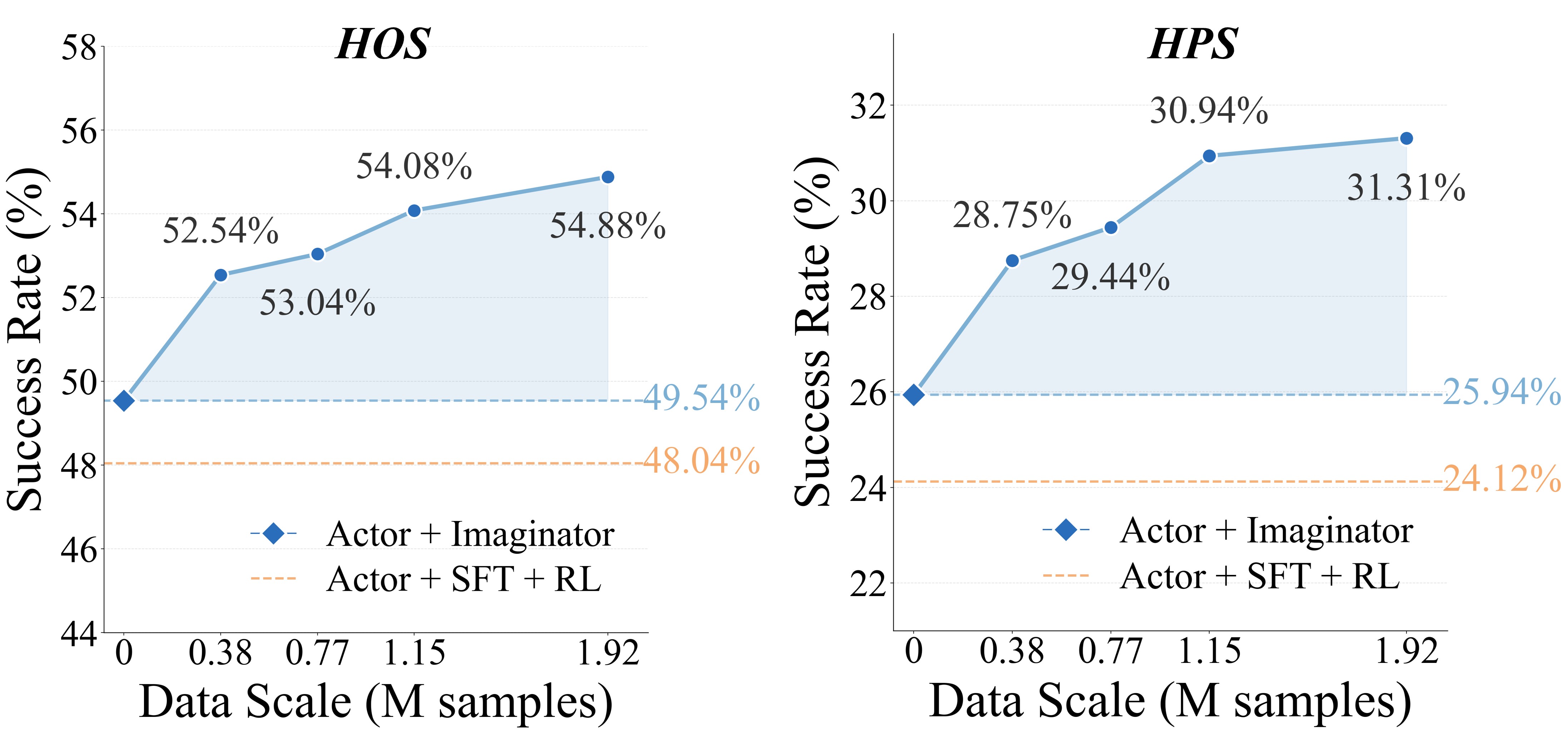}
    \vspace{-3mm}
    \caption{\textbf{Scaling Laws of Imagination.} We use Qwen2.5-VL-3B as the Actor model.}
    \label{fig:scale}
  \end{minipage}%
  \hfill % 在两个 minipage 之间添加弹性空白，将它们推向两侧
  % 右侧表格区域：占据 52% 的页面宽度
  \begin{minipage}[c]{0.47\textwidth}
    \centering
    \small
    \captionof{table}{\textbf{Ablation on Guidance Type.} Evaluated on (\hos) with HVS-3B as Actor Model, where \textit{layout} represents the whole-scene layout modeling during imagination and $T$=0 means without probabilistic sampling.}
    \vspace{-2mm}
    \label{tab:ablation-hos}
    \setlength{\tabcolsep}{0.5mm}{\scalebox{0.75}{
    \begin{tabular}{lccccc}
    \toprule
    \textbf{Imagination Guidance} & \textbf{Overall} & Easy & Medium & Hard \\
    \midrule
    Actor          & 48.04 & 61.79 & 26.14 & 45.18 \\
    Actor + Random Suggestions        & 20.83 & 23.09 & 19.32 & 20.14 \\
    Actor + Pixel Imagination  & 46.17  & 61.46 & 27.27 &  42.39  \\
    Actor + Imaginator (\textit{w.o. layout})     & 41.00 & 59.84 & 34.09 & 34.56 \\
    Actor + Imaginator (\textit{T}=0)                & 61.54 & 74.47 & 58.52 & 56.93 \\
    \midrule
    Actor + Imaginator (\textit{full}) & \textbf{62.75} & \textbf{75.77} & \textbf{64.20} & \textbf{57.61} \\
    \bottomrule
    \end{tabular}
    }}
  \end{minipage}
  \vspace{-4mm}
\end{figure}

\begin{figure}[t]
  \centering
  \includegraphics[width=0.99\textwidth]{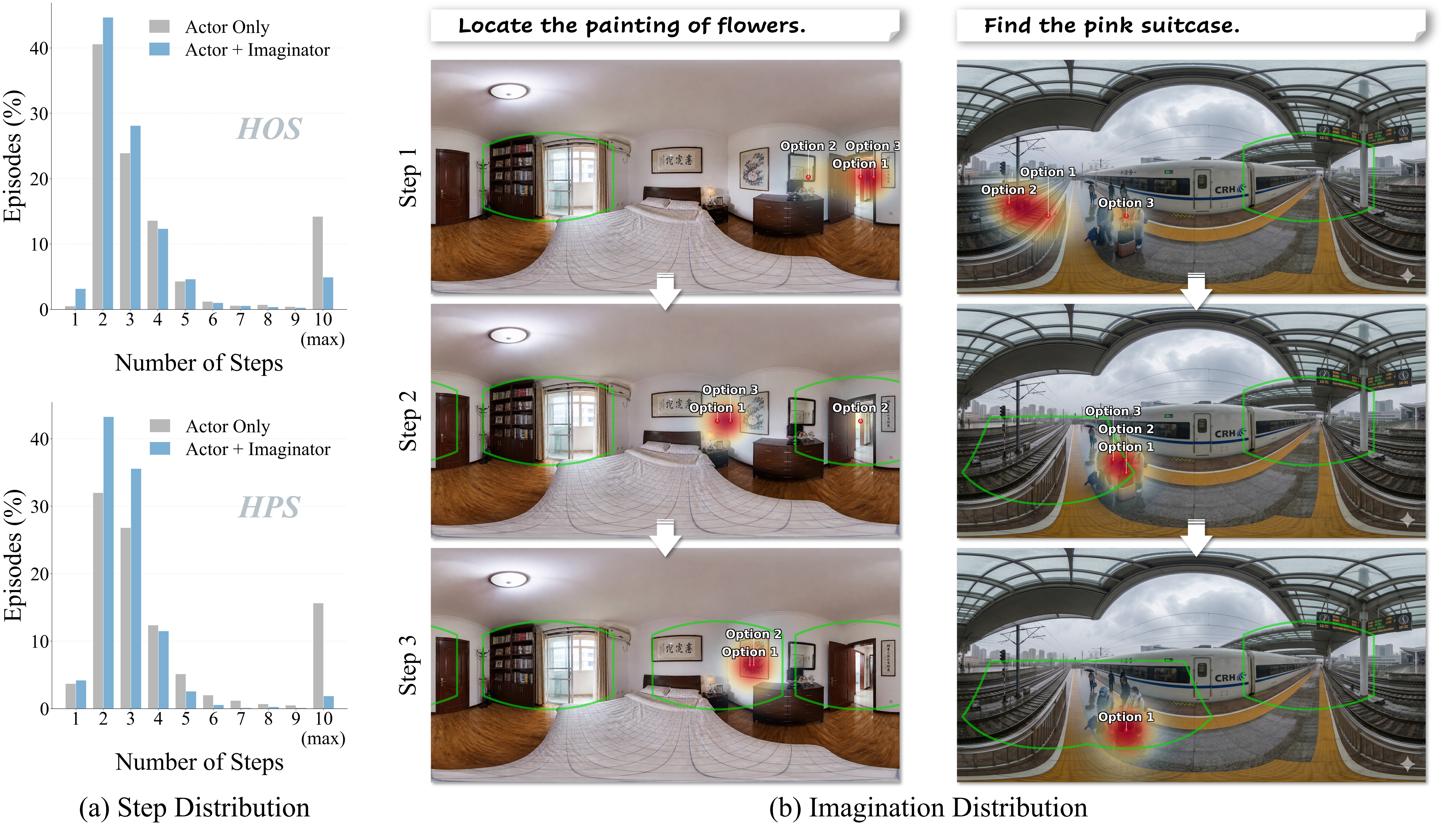}
  \vspace{-2mm}
  \caption{\textbf{Distribution Analysis.} (a) Histograms of the step distribution, illustrating the percentage of search episodes that terminate at each step for both the baseline (Actor Only) and our joint pipeline across the \hos\ and \hps\ tasks. (b) Step-by-step heatmap visualizations of the Imaginator's probabilistic spatial suggestions. The heat areas indicate the probability distribution of the imagined target coordinates, which are projected onto the complete 360$^\circ$ environment.}
  \vspace{-2mm}
  \label{fig:distributions}
\end{figure}

\begin{figure}[t]
  \centering
  \includegraphics[width=\textwidth]{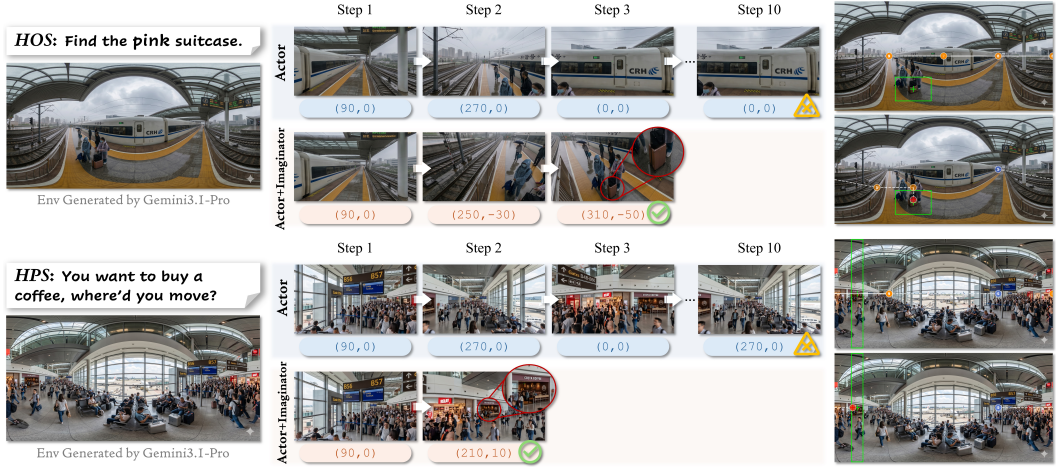}
  \caption{\textbf{Qualitative Search Trajectories in the wild.} Evaluated in complex environments generated by Gemini 3.1-Pro. The Actor loses context and falls into an infinite dead loop, whereas our decoupled framework leverages layout imagination to locate the target within 2-3 efficient steps.}
  \label{fig:vis_wild}
\end{figure}

\subsection{Ablation Studies}
\label{subsec:ablations}

\paragraph{Necessity of Structured Layouts and Probabilistic Sampling.} 
Table~\ref{tab:ablation-hos} dissects the specific components of our imagination guidance using the HVS-3B Actor. First, injecting random suggestions severely degrades performance (20.83\%), confirming the Actor's reliance on external priors. To evaluate alternative representation spaces, we implement a \textit{pixel imagination} baseline by fine-tuning FLUX.1-Fill~\cite{labs2025flux} on $\sim$1.4M panoramas. At each step, FLUX out-paints the unobserved regions conditioned on the accumulated NFoVs, and the generated panorama is fed into the Actor. However, modeling low-level pixel noise proves sub-optimal for high-level reasoning, yielding inferior navigation signals (46.17\%). Furthermore, ablating the structured \texttt{[Observed]} and \texttt{[Imagined]} context (\textit{w.o. layout}) drops the success rate to 41.00\%, lagging even behind the unaugmented baseline Actor (48.04\%). This confirms that explicitly modeling global spatial relationships is essential to prevent hallucinated priors. Finally, our probabilistic sampling (\textit{full}, 62.75\%) consistently outperforms deterministic greedy decoding (\textit{T}=0, 61.54\%).

\paragraph{Data Scaling Laws.} 
Fig.~\ref{fig:scale} demonstrates the scaling dynamics of the \textbf{Imaginator}. Increasing its pseudo-pretrained data from 0 to 1.92M samples yields monotonic improvements across both \hos\ and \hps\ tasks. Notably, equipping the Actor with an Imaginator trained on merely 0.38M samples is already sufficient to surpass the costly RL-finetuned baseline (orange dashed line). We also discover evaluating the Imaginator in isolation yields modest results (\hos\ 37.62\%, \hps\ 25.31\%), as its single-step sampling lacks sequential reasoning. Yet, integrating it with the Actor unlocks massive gains, proving our framework is not a trivial ensemble, but a complementary architecture that grounds generative spatial priors with logical action planning.

\paragraph{Search Efficiency and Distribution Analysis}
Beyond success rates, spatial priors dramatically improve search efficiency. As shown in Fig.~\ref{fig:distributions}(a), the baseline Actor frequently tails toward the 10-step failure cutoff due to redundant blind rotations. Conversely, our joint pipeline shifts the distribution sharply leftward, with most searches rapidly converging within 1-3 steps. Compared to pixel-level imagination (e.g., FLUX.1), which incurs a prohibitive latency of 20-30s per step, a single Imaginator forward pass requires only $\sim$2.5s. Even when sampling multiple spatial suggestions, the total latency remains constrained to 5-7s, drastically reducing inference costs. Fig.~\ref{fig:distributions}(b) visualizes this underlying mechanism. Initially, given a limited field-of-view, the Imaginator generates a broad but logical distribution of spatial hypotheses to guide early exploration. Crucially, as the agent rotates and accumulates visual evidence, the environmental uncertainty rapidly decreases. The predictive heatmap progressively narrows step-by-step, seamlessly converging onto the exact target location and completely eliminating the need for aimless 360$^\circ$ sweeping.

\subsection{Qualitative Results in the Wild}
\label{subsec:qualitative}

To validate real-world robustness, we evaluate the framework on complex, out-of-distribution panoramas generated by Gemini 3.1-Pro (Fig.~\ref{fig:vis_wild}). In the \hos\ scenario (train platform), the baseline Actor is trapped in an oscillating loop and fails. Conversely, the Imaginator instantly models the longitudinal layout, guiding the Actor to foveate the target by Step 3. Similarly, in the \hps\ scenario (airport terminal), the joint pipeline imagines occluded storefronts behind the crowd to locate the coffee shop pathway in just 2 steps. Additional qualitative results are provided in the supplementary material.

\section{Conclusion and Limitations}
\label{sec:conclusion}

In this paper, we introduced \textit{Imagining in 360$^\circ$}, a decoupled framework for Humanoid Visual Search (\hvs) that separates exploration into a generative Imaginator and an executive Actor. This mitigates the cognitive and data-scalability bottlenecks of monolithic models. By leveraging masked world modeling, the Imaginator provides probabilistic semantic layouts that enable the Actor to hedge against uncertainty and conduct efficient, directed searches. Experiments validate that the Imaginator universally boosts foundational MLLMs on \hvs tasks.
Despite its success, our framework relies on an open-loop execution paradigm where the Actor cannot provide feedback on the Imaginator's generated priors. Future work will explore incorporating a \textit{Critic} module to dynamically evaluate and correct hypotheses, enabling a fully closed-loop perception-action-evaluation workflow.

{
% \newpage
    \small
    \bibliographystyle{ieeenat_fullname}
    \bibliography{main}
}

%%%%%%%%%%%%%%%%%%%%%%%%%%%%%%%%%%%%%%%%%%%%%%%%%%%%%%%%%%%%

\appendix

\section{Technical Appendices and Supplementary Material}
% Technical appendices with additional results, figures, graphs, and proofs may be submitted with the paper submission before the full submission deadline (see above). You can upload a ZIP file for videos or code, but do not upload a separate PDF file for the appendix. There is no page limit for the technical appendices. 

% Note: Think of the appendix as ``optional reading'' for reviewers. The paper must be able to stand alone without the appendix; for example, adding critical experiments that support the main claims to an appendix is inappropriate. 

\subsection{More Details on Data Curation}
\label{subsec:supp_data}

\paragraph{Dataset Distributions.}

To complement the main text, we provide a comprehensive breakdown of our automated data curation pipeline, the cleaning process for supervised fine-tuning (SFT), and the statistical distribution of the curated dataset.

\paragraph{Automated Pseudo-Label Pipeline.}
As discussed in the main paper, our massive training corpus of $\sim$1.92M samples is constructed without any manual trajectory annotations. The automated pipeline consists of three stages:
\begin{enumerate}[leftmargin=*, nosep]
    \item \textbf{Scene Filter}: We utilize Qwen3-VL-235B~\cite{qwen3} to classify and filter a mixed pool of unlabeled panoramas (including Sun360~\cite{xiao2012recognizing}, Matterport3D~\cite{chang2017matterport3d}, DiT360 generated~\cite{feng2025dit360}). Approximately 20k high-quality panoramas are retained across indoor, transit, and urban categories.
    \item \textbf{Detection Labels}: The teacher model proposes $\sim$20 semantically salient targets per scene, localizing each to an $(x, y)$ position in equirectangular coordinates. After token-level perplexity filtering and spatial deduplication, we obtain 5-15 high-confidence semantic anchors per panorama.
    \item \textbf{Trajectory Expansion}: For each labeled panorama, we select 3 salient targets and synthetically generate 24 narrow field-of-view (NFoV) trajectories (rendered at $960\times720$, FOV=$100^\circ$). Crucially, half of these (12 trajectories) are ``target-avoiding'' paths specifically designed so that all views remain distant from the target. This forces the model to imagine target locations from non-overlapping views. This deterministic expansion yields the final $\sim$1.92M single-step \texttt{[Observed]}/\texttt{[Imagined]} SFT samples.
\end{enumerate}

\begin{figure}[t]
  \centering
  \includegraphics[width=1.0\textwidth]{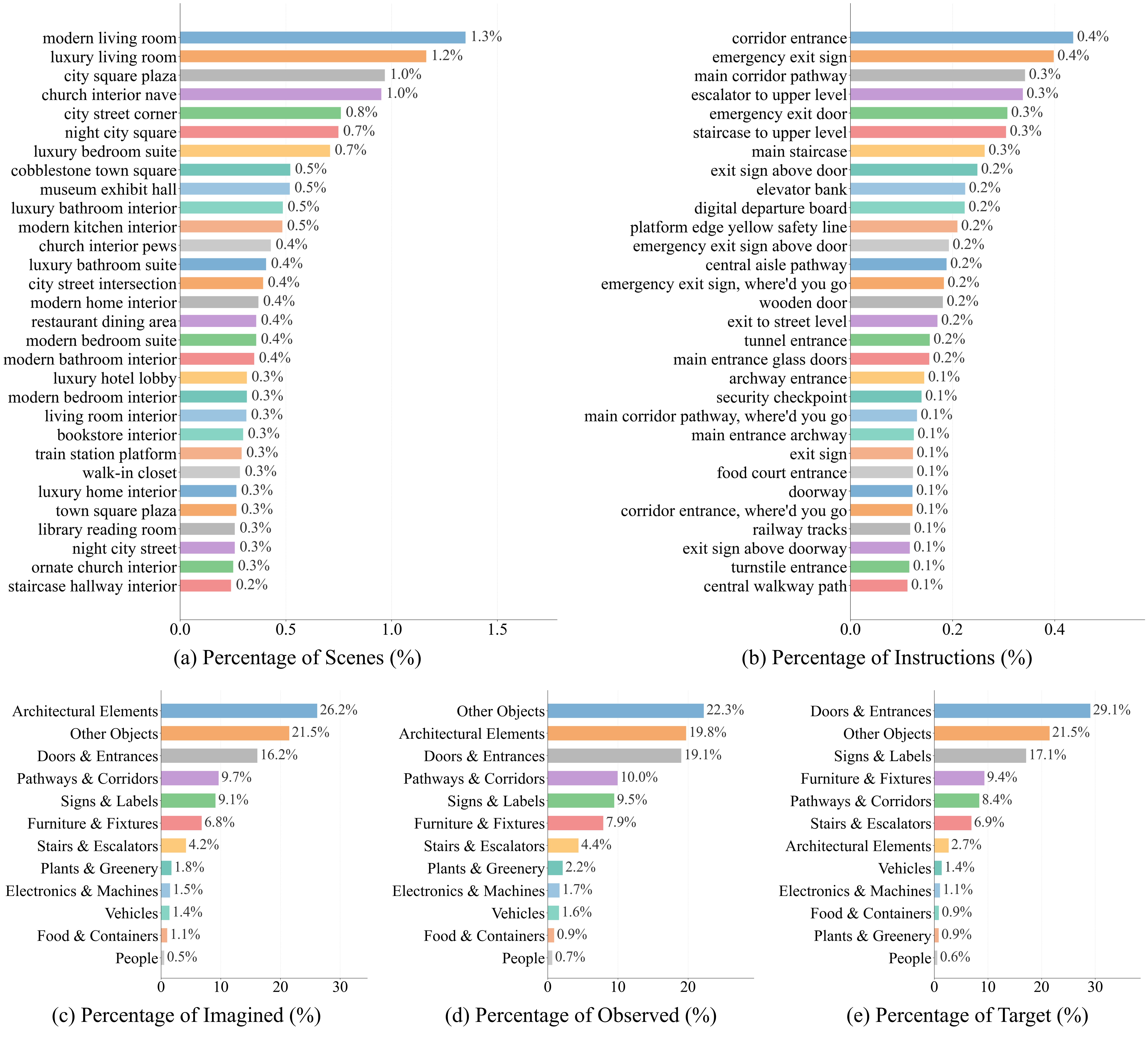}
  \caption{\textbf{Overview of the Curated Training Dataset.} (a) Fine-grained scene category distribution. (b) Distribution of target-search instructions. (c-e) Semantic category distributions for Imagined entities, Observed entities, and Search Targets, respectively.}
  \label{fig:data_supp}
\end{figure}

\paragraph{SFT Data Cleaning.}
For the final task alignment stage, we aggressively filter the raw H*Bench~\cite{yu2025thinking} SFT annotations to ensure high-quality reasoning. We enforce a strict structural requirement: a training sample is retained if and only if its response contains both an \texttt{[Observed]} and an \texttt{[Imagined]} block, since some of the cases lack raw panoramas, so the intermediate structured layouts are also absent and replies degenerate into a simple spatial guess (e.g., just outputting \texttt{suggest check(yaw, pitch)}), as they would teach the model to guess rather than explicitly model the space. This rigorous cleaning process retains 2,100 HOS samples (73.8\% retention) and 2,148 HPS samples (81.0\% retention), forming a highly curated alignment corpus of 4,248 samples.

Figure~\ref{fig:data_supp} illustrates the diversity and semantic richness of our curated dataset. As shown in Fig.~\ref{fig:data_supp}(a) and (b), the dataset covers a highly diverse distribution of environments ranging from indoor domains (e.g., modern living rooms, luxury bedroom suites) to outdoor areas (e.g., city square plazas), accommodating a wide array of corresponding navigation instructions. Furthermore, Fig.~\ref{fig:data_supp}(c-e) detail the semantic composition of the spatial layouts. Notably, ``Architectural Elements'' (26.2\%) and ``Doors \& Entrances'' (16.2\%) dominate the \texttt{[Imagined]} categories (Fig.~\ref{fig:data_supp}c). This inherently aligns with the cognitive requirements of our task: architectural features and doors serve as critical structural anchors for extrapolating the layout of unseen regions. Similarly, ``Doors \& Entrances'' and ``Objects'' constitute large portions (29.1\% and 21.5\%) of the search targets (Fig.~\ref{fig:data_supp}e), accurately reflecting the object/path-finding and navigational nature of practical \hvs applications.

\subsection{More Details on Training and Inference}
\label{subsec:supp_training}

In this section, we provide extended implementation details regarding the two-stage training of the Imaginator, the pixel imagination baseline model, and the inference-time sampling strategy.

\paragraph{Imaginator Optimization.}
The Imaginator is initialized from {Qwen3-VL-8B-Instruct}~\cite{qwen3} and optimized via full-parameter supervised fine-tuning (SFT). We utilize the AdamW optimizer with a learning rate of $1 \times 10^{-5}$, a cosine decay schedule, and a 3\% linear warmup. To ensure training stability and memory efficiency, we employ DeepSpeed~\cite{rasley2020deepspeed} ZeRO~\cite{rajbhandari2020zero} Stage 2 with gradient sharding and BF16 mixed precision. 
\begin{itemize}[leftmargin=*, nosep]
    \item \textbf{Stage 1 (Pseudo-pretraining)}: The model is trained on the $\sim$1.92M synthetic sample corpus for 1 epoch. We use a global batch size of 192 (8 GPUs $\times$ 6 per-device $\times$ 4 gradient accumulation).
    \item \textbf{Stage 2 (Clean-data SFT)}: The pseudo-pretrained checkpoints are fine-tuned on the curated H*Bench set ($\sim$4,248 samples) for 2 epochs to compensate for the smaller corpus size. The global batch size is adjusted to 8.
\end{itemize}
Input images are resized to a maximum of 921,600 pixels (1280$\times$720), and sequences are truncated at 8,192 tokens, which covers the 99th-percentile length of our structured layouts.

\begin{table}[t]
\centering
\small
\caption{\textbf{Full Results on H*Bench.} We report the Success Rate (\%) across various Vision-Language Model families. `$\times$' denotes the baseline Actor relying solely on its internal CoT. `$\checkmark$' indicates the Actor plugged in with our Imaginator's spatial priors.}

\setlength{\tabcolsep}{4pt}
\setlength{\tabcolsep}{2.3mm}{\scalebox{0.75}{
\begin{tabular}{llcccccccccc}
\toprule
\multicolumn{2}{c}{Actor} & \multirow{2}{*}{Imaginator} & \multicolumn{4}{c}{Humanoid Object Search} & \multicolumn{5}{c}{Humanoid Path Search} \\
\cmidrule(lr){1-2} \cmidrule(lr){4-7} \cmidrule(lr){8-12}
Family & Params &  & \textbf{Overall} & Easy & Medium & Hard & \textbf{Overall} & Easy & Medium & Hard & Extreme \\
\midrule
\rowcolor{gray!15} \multicolumn{12}{l}{\textit{Open-Source Models}} \\
\multirow{8}{*}{Qwen2.5-VL} & \multirow{2}{*}{3B} & $\times$ & 21.17 & 18.54 &  16.48 &  22.68  & 7.69 & 9.20 & 9.65 & 3.50 & 8.33 \\
 &  & $\checkmark$ & 49.54 & 59.19 &  51.70 &  45.62 & 25.94 & 34.80 & 22.15 & 24.77 & 15.74 \\
\cmidrule(lr){2-12}
 & \multirow{2}{*}{7B} & $\times$ & 12.79 & 31.22 & 7.39 & 6.34 & 9.88 & 12.20 & 11.18 & 7.24 & 6.94 \\
 &  & $\checkmark$ & 57.42 & 70.73 & 51.70 & 52.95 & 28.38 & 35.20 & 24.34 & 29.44 & 18.98 \\
\cmidrule(lr){2-12}
 & \multirow{2}{*}{32B} & $\times$ & 32.08 & 50.57 & 27.84 & 25.48 & 17.56 & 21.80 & 15.35 & 19.39 & 8.80 \\
 &  & $\checkmark$ & 41.12 & 66.34 & 36.36 & 32.01 & 21.12 & 25.60 & 19.74 & 21.26 & 13.43 \\
\cmidrule(lr){2-12}
 & \multirow{2}{*}{72B} & $\times$ & 29.54 & 36.91 & 21.59 & 27.59 & 12.94 & 17.40 & 9.21 & 15.65 & 5.09 \\
 &  & $\checkmark$ & 62.25 & 70.08 & 60.80 & 59.42 & 38.19 & 45.00 & 36.18 & 39.25 & 24.54 \\
\midrule
\multirow{6}{*}{Qwen3-VL} & \multirow{2}{*}{30B-A3B} & $\times$ & 24.96 & 54.63 & 30.11 & 13.05 & 8.44 & 11.40 & 7.24 & 8.88 & 3.24 \\
 &  & $\checkmark$ & 56.67 & 74.15 & 55.68 & 50.09 & 30.94 & 41.80 & 26.32 & 30.14 & 17.13 \\
\cmidrule(lr){2-12}
 & \multirow{2}{*}{32B} & $\times$ & 34.38 & 66.83 & 36.36 & 21.75 & 23.19 & 26.00 & 24.78 & 22.43 & 14.81 \\
 &  & $\checkmark$ & 57.46 & 73.98 & 51.70 & 51.77 & 31.31 & 37.60 & 30.04 & 30.14 & 21.76 \\
\cmidrule(lr){2-12}
 & \multirow{2}{*}{235B} & $\times$ & 23.62 & 57.24 & 26.70 & 10.44 & 21.44 & 26.80 & 19.96 & 23.36 & 8.33 \\
 &  & $\checkmark$ & 65.04 & 81.46 & 60.23 & 59.29 & 38.25 & 46.60 & 35.31 & 37.15 & 27.31 \\
\midrule
\multirow{4}{*}{Gemma-3} & \multirow{2}{*}{4B} & $\times$ & 18.17 & 41.79 & 25.57 & 8.33 & 14.50 & 17.80 & 16.67 & 10.98 & 9.26 \\
 &  & $\checkmark$ & 57.75 & 72.03 & 60.23 & 52.02 & 35.06 & 43.80 & 27.41 & 36.45 & 28.24 \\
\cmidrule(lr){2-12}
 & \multirow{2}{*}{12B} & $\times$ & 22.88 & 54.96 & 32.39 & 9.57 & 15.31 & 16.80 & 16.23 & 14.49 & 11.57 \\
 &  & $\checkmark$ & 54.62 & 63.90 & 48.86 & 51.71 & 33.56 & 43.00 & 30.04 & 32.24 & 21.76 \\
 \midrule
 \multirow{2}{*}{Kimi-VL} & \multirow{2}{*}{A3B} & $\times$ & 4.33 & 12.36 & 0.57 & 1.68 & 5.62 & 7.20 & 6.80 & 2.34 & 6.02 \\
 &  & $\checkmark$ & 32.46 & 40.00 & 27.84 & 30.08 & 21.00 & 28.20 & 16.89 & 21.96 & 11.11 \\
\midrule
\multirow{4}{*}{InternVL3.5} & \multirow{2}{*}{4B} & $\times$ & 10.00 & 14.80 & 13.07 & 7.83 & 8.62 & 12.80 & 9.43 & 6.31 & 1.85 \\
 &  & $\checkmark$ & 62.42 & 69.59 & 62.50 & 59.66 & 34.75 & 44.80 & 28.95 & 35.51 & 22.22 \\
\cmidrule(lr){2-12}
 & \multirow{2}{*}{8B} & $\times$ & 11.04 & 23.74 & 6.25 & 6.71 & 7.06 & 9.60 & 6.80 & 6.31 & 3.24 \\
 &  & $\checkmark$ & 66.33 & 76.10 & 68.18 & 62.40 & 38.38 & 50.60 & 28.73 & 39.95 & 27.31 \\
\midrule
 \rowcolor{gray!15} \multicolumn{12}{l}{\textit{Proprietary Models}} \\
 \multirow{4}{*}{Gemini-2.5} & \multirow{2}{*}{Flash} & $\times$ & 43.70 & 79.50 & 59.70 & 28.30 & 27.70 & 35.80 & 22.60 & 31.30 & 12.50 \\
 &  & $\checkmark$ & 63.20 & 86.30 & 69.30 & 53.70 & 37.70 & 50.20 & 30.30 & 42.30 & 15.30 \\
 \cmidrule(lr){2-12}
 & \multirow{2}{*}{Pro} & $\times$ & 55.00 & 88.00 & 62.50 & 41.60 &  37.90 & 43.80 & 36.40 & 40.70 & 21.80 \\
 &  & $\checkmark$ & 66.10 & 89.60 & 68.80 & 56.80 & 43.10 & 51.00 & 39.90 & 48.80 & 20.40 \\
 \midrule
 \rowcolor{gray!15} \multicolumn{12}{l}{\textit{Finetuned Models}} \\
 \multirow{2}{*}{HVS} & \multirow{2}{*}{3B} & $\times$ & 48.04 & 61.79 & 26.14 & 45.18 & 24.12 & 36.60 & 17.76 & 22.66 & 11.57 \\
 &  & $\checkmark$ & 62.75 & 75.77 & 64.20 & 57.61 & 39.38 & 49.40 & 32.89 & 41.12 & 26.39 \\
\bottomrule
\end{tabular}
}}
\label{tab:main_results_supp}
\end{table}

\paragraph{Pixel Imagination Baseline.}
With the rapid development of diffusion-based generative models~\cite{lugmayr2022repaint,rombach2022high,zhang2023adding,podell2023sdxl,zhang2025uniser}, effective generative priors are gained by pretraining on billions of images representing the world's knowledge. To evaluate alternative representation spaces in our ablation study, we implemented a pixel-level imagination baseline using a panoramic outpainting model. Built upon \texttt{FLUX.1-Fill-dev}~\cite{labs2025flux}, the model is adapted via LoRA fine-tuning (rank 64, $\alpha=64$) and initialized from the DiT360~\cite{feng2025dit360} generation checkpoint, which provides strong equirectangular geometry priors. At each step, this model out-paints the unobserved regions conditioned on the accumulated NFoV observations to generate a complete panorama. It is optimized using a masked flow-matching MSE loss to ensure gradients flow exclusively through the generated unknown regions:
\begin{equation}
\mathcal{L} = \frac{\sum [ w(\sigma) \cdot \|v_{\text{pred}} - v_{\text{target}}\|^2 \cdot M_{\text{unknown}} ]}{|M_{\text{unknown}}| \cdot C}
\end{equation}
where $M_{\text{unknown}}$ represents the mask of the unobserved regions. Training is conducted for 10,000 steps with an AdamW optimizer ($lr=2\times10^{-5}$) on a multi-source panorama corpus from~\cite{zhang2026mtpano}.

In the inference cycle, once the agent receives a partial panoramic canvas built from current NFoV views (containing both explored and masked unknown regions), the pixel imagination model executes a 30-step diffusion process via the finetuned DiT360 to outpaint the unobserved areas, which are then blended with the original known regions to ensure visual consistency. The HVS agent subsequently perceives both the original partial observation and the completed reference panorama to guide its next action.

\paragraph{Inference-time Sampling Strategy.}
During the Humanoid Visual Search process, the Imaginator captures spatial uncertainty through a temperature-decay sampling schedule. At each step $t$, we sample $N$ spatial hypotheses where the temperature $T_t$ follows:
\begin{equation}
T_t = T_1 \cdot \gamma^{(t-1)}
\end{equation}
We set the initial temperature $T_1 = 0.7$ and the decay factor $\gamma = 0.85$. This decay schedule ensures that the agent explores diverse spatial possibilities in the early stages of search (high $T$) and transitions to focused exploitation (low $T$) as visual evidence accumulates. All suggestions are filtered via top-$k$ logit sampling ($k=50$) to maintain semantic coherence. We use this temperature sampling strategy in the first 3 steps since uncertain search mostly happens at the beginning stage, with the observation gaining, the uncertainty quickly drops and converges to the target.

\begin{figure}[t]
  \centering
  \includegraphics[width=1.0\textwidth]{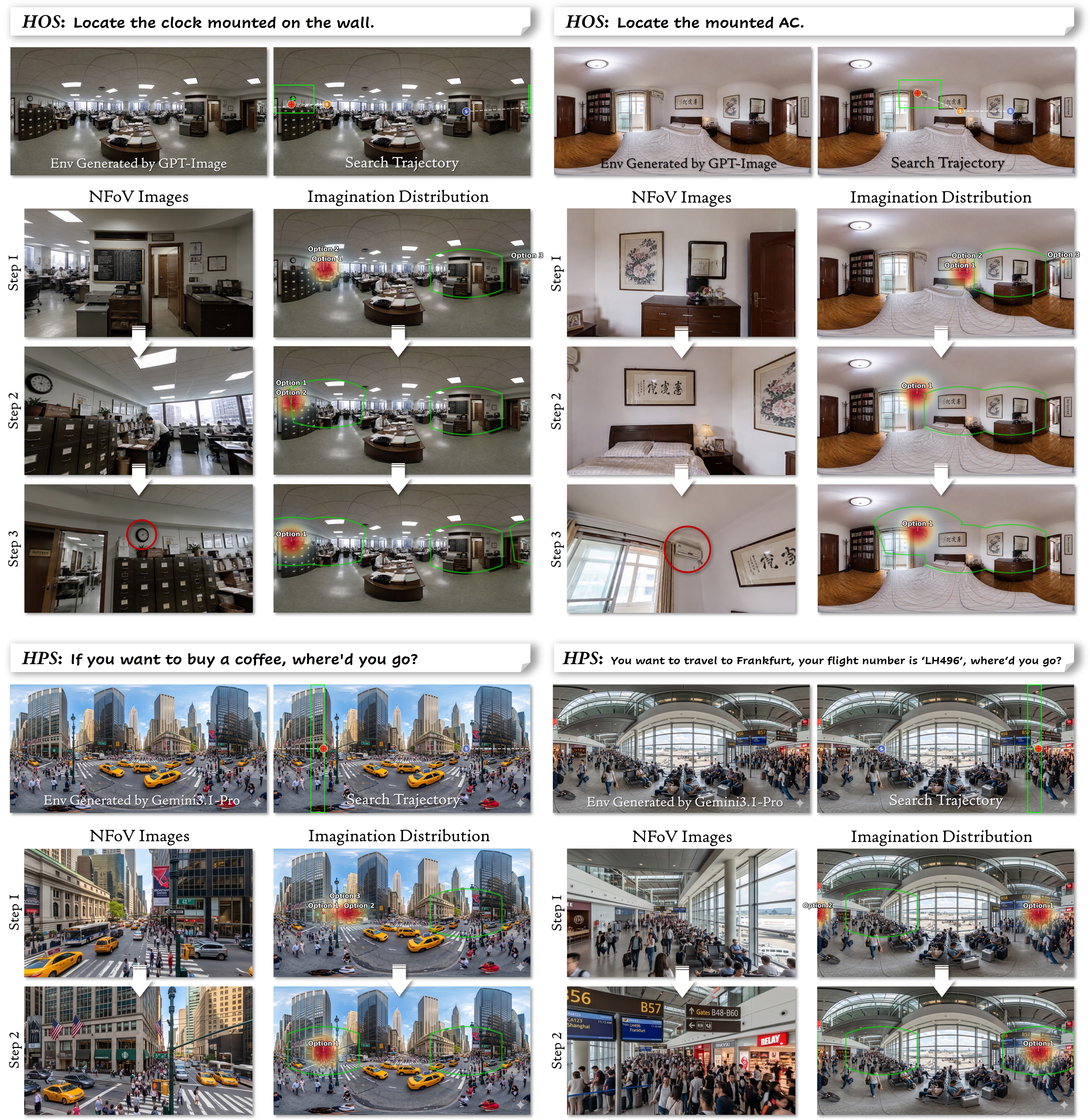}
  \caption{\textbf{Qualitative search trajectories in synthetic panoramas.} Evaluated in complex indoor and outdoor environments generated by GPT-Image and Gemini 3.1-Pro. The step-by-step imagination distributions (heatmaps) demonstrate how the framework progressively refines its spatial hypotheses as new visual evidence is gathered, effectively guiding both HOS and HPS tasks.}
  \label{fig:vis_more_syn}
\end{figure}

\begin{figure}[t]
  \centering
  \includegraphics[width=1.0\textwidth]{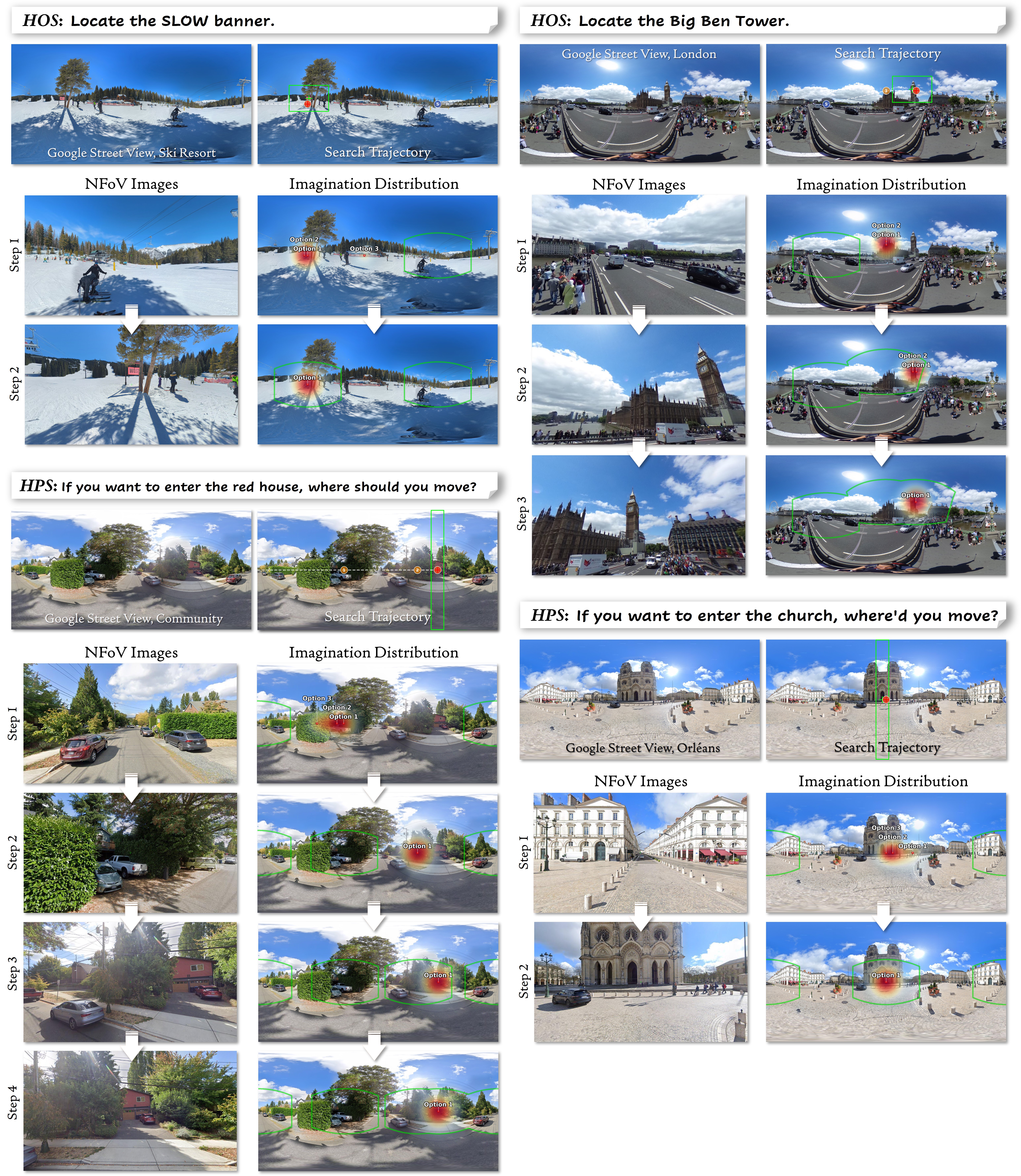}
  \caption{\textbf{Qualitative search trajectories in real-world panoramas.} Evaluated in diverse, in-the-wild Google Street View environments. The decoupled pipeline demonstrates strong zero-shot robustness in unstructured outdoor scenes (e.g., ski resorts, bustling city streets, and suburban neighborhoods), efficiently narrowing down target locations.}
  \label{fig:vis_more_real}
\end{figure}

\subsection{More Quantitative Comparisons}
\label{subsec:supp_quantitative}

To provide a comprehensive evaluation of our proposed framework, Table~\ref{tab:main_results_supp} details the full quantitative results on the H*Bench~\cite{yu2025thinking} dataset, expanding upon the main paper by including a wider array of model families (such as InternVL3.5~\cite{chen2025internvl35advancingopensource} and Qwen3-VL-30B-A3B) and breaking down the success rates across all difficulty tiers (Easy, Medium, Hard, and Extreme). 

\paragraph{Universal Boosting Across Model Scales.}
The most prominent takeaway from Table~\ref{tab:main_results_supp} is the universal applicability of the Imaginator's spatial priors. As a plug-and-play context provider, it drastically enhances the search capabilities of open-source models regardless of their parameter size. For instance, the InternVL3.5 family exhibits staggering improvements: the 8B model jumps from a mere 11.04\% to 66.33\% on Humanoid Object Search (HOS), and from 7.06\% to 38.38\% on Humanoid Path Search (HPS). Similar transformative gains are observed across the Qwen2.5-VL, Qwen3-VL, and Gemma-3 families. This proves that explicit 3D spatial modeling effectively bridges the capability gap for smaller or less embodied-aligned models.

\paragraph{Efficacy on State-of-the-Art Proprietary Models.}
While cutting-edge proprietary models like Gemini-2.5 Pro exhibit strong baseline performance relying on their massive internal world knowledge (e.g., 55.00\% on HOS), they are not immune to the cognitive burdens of monolithic 360$^\circ$ reasoning. By integrating our Imaginator, Gemini-2.5 Pro reaches 66.10\% on HOS and 43.10\% on HPS. Furthermore, Gemini-2.5 Flash sees a massive +19.5\% absolute improvement on HOS (43.70\% $\rightarrow$ 63.20\%), highlighting that explicit spatial hypotheses are crucial even for highly advanced multimodal systems.

\paragraph{Surpassing Task-Specific RL Models.}
Our framework's superiority is further solidified when evaluated alongside HVS-3B, a model specifically fine-tuned with Reinforcement Learning (RL) for this exact task. A standalone HVS-3B achieves 48.04\% on HOS. However, when deployed as the Actor within our decoupled pipeline, its performance surges to 62.75\%. This indicates that forcing a single model to simultaneously imagine layouts and plan actions imposes a fundamental ceiling—one that can be broken by explicitly decoupling the probabilistic spatial priors.

\paragraph{Robustness in Extreme Environments.}
The difficulty breakdown in Table~\ref{tab:main_results_supp} reveals that the baseline Actor models degrade rapidly as the environments become more complex (Hard and Extreme subsets). Without the Imaginator, most open-source models fall below 10\% success rate in Extreme HPS scenarios, often succumbing to blind 360$^\circ$ sweeping. In contrast, the joint pipeline maintains remarkable robustness, frequently tripling or quadrupling the success rates in these highly occluded and challenging conditions.

\subsection{More Qualitative Visualizations}
\label{subsec:supp_vis}

To further demonstrate the robustness and generalizability of our decoupled framework, we provide additional step-by-step search trajectories in highly complex, out-of-distribution (OOD) environments across both Humanoid Object Search (HOS) and Humanoid Path Search (HPS) tasks. 

\paragraph{Search in Synthetic Environments.}
Figure~\ref{fig:vis_more_syn} showcases the agent's performance in panoramic environments synthesized by state-of-the-art generative models, including GPT-Image and Gemini 3.1-Pro. These scenes feature densely packed indoor rooms (e.g., offices with cluttered desks) and busy outdoor hubs (e.g., city intersections, airport terminals). Despite the unconventional layouts or minor artifacts inherent to AI-generated images, the Imaginator successfully captures the structural semantics of the unobserved regions. As the agent accumulates Narrow Field-of-View (NFoV) observations step-by-step, the imagination distribution (shown as predictive heatmaps) progressively narrows, eliminating spatial uncertainty and efficiently guiding the Actor toward the targets (e.g., locating a wall-mounted clock or the pathway to flight LH496).

\paragraph{Search in Real-World Environments.}
Figure~\ref{fig:vis_more_real} illustrates search trajectories in unstructured, real-world panoramas sampled directly from Google Street View. These environments present significant navigational challenges due to their vast scales, dynamic elements (e.g., pedestrians, skiers), and extreme topological diversity, ranging from snowy mountain resorts to historic European plazas. Nevertheless, the Imaginator consistently generates logical spatial priors. For instance, when tasked with locating the ``Big Ben Tower'' or determining the path to ``enter the church,'' the model effectively leverages early architectural cues to hypothesize the target's location. This allows the agent to lock onto the correct trajectory within just 2 to 4 steps, completely avoiding redundant 360$^\circ$ sweeping.

\subsection{Social Impacts}
\label{subsec:social_impacts}

The proposed framework for Humanoid Visual Search (\hvs) has significant potential for positive social impact, particularly in enhancing the autonomy of service robots and search-and-rescue agents in complex, unstructured environments. By improving search efficiency and robustness, our approach can facilitate more reliable assistance in household chores, elderly care, and emergency response, ultimately improving quality of life and operational safety. 

However, we also acknowledge potential ethical considerations. The deployment of active 360$^\circ$ perception systems may raise privacy concerns regarding data collection in residential or public spaces. Furthermore, like any advanced vision-action system, there is a risk of misuse in unauthorized surveillance applications. We encourage the research community to prioritize privacy-preserving mechanisms and adhere to strict ethical guidelines in future embodied AI deployments.

\subsection{Acknowledgement}
The authors want to thank Zekai Shao for his constructive suggestions.

%%%%%%%%%%%%%%%%%%%%%%%%%%%%%%%%%%%%%%%%%%%%%%%%%%%%%%%%%%%%

% \newpage
% \input{checklist.tex}

\end{document}